\documentclass[sigconf, nonacm, pbalance]{acmart}
\usepackage{soul}
\usepackage{threeparttable}
\usepackage[ruled,vlined,linesnumbered]{algorithm2e}
\usepackage{multirow}   
\usepackage{svg}
\usepackage{array}
\usepackage{booktabs}
\usepackage{subcaption}
\usepackage{siunitx}
\usepackage{tabularx}
\usepackage[inline,shortlabels]{enumitem}

\graphicspath{{./figures/}{./figures/results}} 

\usepackage{cleveref} 
\crefname{section}{§\!}{§§\!}
\Crefname{section}{Section}{Sections}
\crefname{figure}{Fig.}{Figs.}
\Crefname{figure}{Figure}{Figures}
\crefname{table}{Table}{Tables}
\Crefname{table}{Table}{Tables}
\crefname{algorithm}{Algorithm}{Algorithms}
\Crefname{algorithm}{Algorithm}{Algorithms}

\newcommand{\eg}{e.g., }
\newcommand{\ie}{i.e., }

\def\trackchanges{0} 
\ifthenelse{\equal{\trackchanges}{1}}{
    \newcommand{\newtext}[1]{\textcolor{blue}{#1}}
    \newcommand{\oldtext}[1]{\textcolor{purple}{\st{#1}}}
}{
    \newcommand{\newtext}[1]{{#1}}
    \newcommand{\oldtext}[1]{}
} 
\newcommand{\replacetext}[2]{\oldtext{#1}\newtext{#2}}

\AtBeginDocument{%
  }

\begin{document}

\title[Robustness of STGNNs for Fault Location in Partially Observable Grids]{Robustness of Spatio-temporal Graph Neural Networks for Fault Location in Partially Observable Distribution Grids}

\author{Burak Karabulut}
\affiliation{%
  \institution{Ghent University -- imec}
  \city{Ghent}
  \country{Belgium}
}
\email{burak.karabulut@ugent.be}

\author{Carlo Manna}
\affiliation{%
  \institution{VITO}
  \city{Mol}
  \country{Belgium}
}
\email{carlo.manna@vito.be}

\author{Chris Develder}
\affiliation{%
  \institution{Ghent University -- imec}
  \city{Ghent}
  \country{Belgium}
}
\email{chris.develder@ugent.be}

\renewcommand{\shortauthors}{B.\ Karabulut et al.}

\begin{abstract}
Fault location in distribution grids is critical for reliability and minimizing outage durations.
Yet, it remains challenging due to partial observability, given sparse measurement infrastructure.
Recent works show promising results by combining Recurrent Neural Networks (RNNs) and Graph Neural Networks (GNNs) for spatio-temporal learning. 
Still, many modern GNN architectures remain untested for this grid application, while existing GNN solutions have not explored GNN topology definitions beyond simply adopting the full grid topology to construct the GNN graph. We address these gaps by (i)~systematically comparing a newly proposed graph-forming strategy (\emph{measured-only}) to the traditional \emph{full-topology} approach, and 
(ii)~introducing STGNN (Spatio-temporal GNN) models based on GraphSAGE and an improved Graph Attention (GATv2), for distribution grid fault location; (iii)~benchmarking them against state-of-the-art STGNN and RNN baselines on the IEEE~123-bus feeder.
In our experiments, all evaluated STGNN variants achieve high performance and consistently outperform a pure RNN baseline, with improvements  up to $+$11 percentage points F1. Among STGNN models, the newly explored RGATv2 and RGSAGE achieve only marginally higher F1 scores. Still, STGNNs demonstrate superior stability, with tight confidence intervals (within $\pm 1.4\%$) compared to the RNN baseline (up to $\pm 7.5\%$) across different experiment runs. Finally, our proposed reduced GNN topology (\emph{measured-only}) shows clear benefits in both (i)~model training time (6-fold reduction) and (ii)~model performance (up to $+$11 points F1).
This suggests that \emph{measured-only} graphs offer a more practical, efficient, and robust framework for partially observable distribution grids.
\end{abstract}




\maketitle
\section{Introduction}
\label{sec:intro}
Fault location in power distribution networks is essential for maintaining reliability and minimizing outage durations~\cite{ieee1366}. By pinpointing the faulted component\,---\,typically resulting from short circuits caused by weather, equipment failures, or insulation breakdowns~\cite{grainger1994power, teixeira2021fault}\,---\,operators can selectively isolate the affected area and restore service more quickly. Given that modern distribution networks are becoming larger, more interconnected, and increasingly complex\,---\,with the integration of distributed energy resources (DERs) and the electrification of demand (e.g., electric vehicle charging)~\cite{iea2023grids}\,---\,
the importance of accurate and timely fault location 
only continues to grow. Yet, to limit the incurred cost,
distribution networks are typically equipped with only a limited number of measurement devices~\cite{momoh2012smart}, such as micro-phasor measurement units ($\upmu$PMUs). This infrastructural constraint results in limited observability, and motivates the development of fault location methods that can work effectively with sparse measurement data.

Traditional model-based approaches often struggle in these environments due to their reliance on high feeder observability and static assumptions that regarding feeder topology, operating conditions and fault characteristics~\cite{rezapour2023review}. While data-driven methods such as Convolutional Neural Networks (CNNs) and Recurrent Neural Networks (RNNs) offer better flexibility they often lack topological awareness (\ie they cannot fully represent the non-uniform electrical connectivity between buses)~\cite{chen2020fault}.
To more effectively leverage the topological relations inherent in distribution grids, graph-based learning approaches such as Graph Neural Networks (GNNs)~\cite{gori2005new}, have seen increasing attention for fault location; however, several research gaps remain unaddressed: 
\begin{enumerate*}[(i)]
  \item current GNN approaches typically assume high observability via dense 3-phase sensor placement and have not explored alternate graph representations beyond the 1-to-1 mapping of the GNN graph to the actual feeder bus topology; 
  further
  \item there has not been a rigorous performance comparison of the various recent GNN architectures for fault location; and finally,
  \item most prior works typically rely on relatively small, simplified datasets, leaving a gap in validating GNN performance across the diverse load variations and fault scenarios.
\end{enumerate*}
Our study addresses these issues, in particular by
\begin{enumerate}[(1)]
    \item Proposing and systematically assessing a GNN graph construction algorithm (see \cref{sec:logical-topology}) that accounts for partial observability by using only measured buses while reflecting the underlying physical connectivity, and contrasting its performance against the GNN graphs constructed using the full physical topology; 
    \item Proposing STGNN architectures (see \cref{subsec:sfe_gnn}) that remained unexplored for power system fault location in distribution grids, \ie the RGSAGE and improved RGAT models; 
    \item Conducting a systematic quantitative benchmarking study (see \cref{sec:rnn-vs-rgnn}) against state-of-the-art graph models (RGCN) as well as non-GNN baselines (GRU), using a comprehensive dataset of RMS voltage measurements collected via a heterogeneous and sparse placement of 1-, 2-, and 3-phase $\upmu$PMUs. (see \cref{sec:exp}). 
\end{enumerate}
Before discussing the setup of our experiments (\cref{sec:exp}) and their outcomes (\cref{sec:res}), next we first present extensive literature review (\cref{sec:related}), then introduce our proposed methodology (\cref{sec:method}). In this paper's conclusion (\cref{sec:conclusion}) we will finally discuss next steps to extend the current work towards more advanced
fault diagnosis applications. 

\section{Related Work}
\label{sec:related}
The literature on fault location in power distribution systems~\cite{rezapour2023review, chen2016fault} are typically characterized as either model-based or data-driven approaches. 
\textit{Model-based methods} include impedance-based~\cite{chandran2024extended, krishnathevar2012generalized} and voltage sag-based analysis~\cite{turizo2022voltage, buzo2021new}, automated outage mapping~\cite{numair2023fault}, and traveling wave-based techniques~\cite{liu2023novel, thomas2003fault}. 
Despite their effectiveness in traditional radial settings, these methods often rely on extensive measurement infrastructure to ensure sufficient observability, alongside the static assumptions regarding feeder topology, operating conditions and fault characteristics~\cite{ojha2018optimization}. Specifically, while traveling wave methods are often hindered by the high capital costs of required high-speed measurement infrastructure, impedance and sag-based strategies are highly sensitive to static topological assumptions. 
In modern grids, such assumptions can be frequently challenged by the dynamic and evolving operating conditions aforementioned, leading to increased modeling errors and false alarms~\cite{rezapour2023review, mansourlakouraj2021application, chen2020fault}.

Early \textit{data-driven methods} have been thoroughly investigated to address the shortcomings of conventional model-based approaches, e.g.,~\cite{livani2013faulty,livani2013FaultClass, hosseini2018ami}. Traditional machine learning solutions include k-nearest neighbors (kNN), random forests, decision trees, and support vector machines (SVMs), were applied to locate faults by relying on hand-crafted features~\cite{chen2016fault}.
More recent deep learning approaches, such as 
CNNs and RNNs, have shown potential due to their ability to extract patterns from raw or minimally processed measurements.
While CNNs have been particularly effective in identifying spatial relationships, RNNs (e.g., Gated Recurrent Unit (GRU), Long Short-Term Memory (LSTM)) exploit temporal patterns in sequential data~\cite{alzubaidi2021review, nguyen2023spatial}.
Still, these data-driven methods often lack topological awareness (\ie they cannot fully represent the non-uniform electrical connectivity between buses)~\cite{chen2020fault}.
This lack of structural insight may limit their location performance. 

GNNs have seen increasing attention for fault location due to their ability to leverage the topological relations inherent in power networks. Originally proposed by Gori et al.~\cite{gori2005new, scarselli2009graph}, GNNs are specifically designed to process graph-structured data, which naturally translates to power systems, as network components such as buses and their interconnecting distribution power lines  can be modeled as graph nodes and edges. Through message passing, GNNs aggregate information from neighboring busses to capture spatial dependencies across the feeder~\cite{liao2022review}.

Specifically for our considered fault location problem, Chen et al.~\cite{chen2020fault} use \emph{Spectral Graph Convolutional Networks (GCNs)} which perform convolutions via the feeder’s Laplacian matrix in the graph Fourier domain, thus enabling capturing
global structural information across all buses. Still, while effective at leveraging global information, spectral GNNs have traditionally faced scalability challenges in power systems because they require a global mapping of the entire network's structure. Although polynomial approximations \cite{hammond2011wavelets, defferrard2016convolutional} have significantly reduced this computational overhead, these methods remain tied to a fixed global basis. Consequently, while data augmentation can provide a degree of stability against unseen reconfigurations~\cite{chen2020fault}, significant topological changes still require a realignment of these global filters to match the new grid structure, making them theoretically less flexible for highly dynamic environments~\cite{liao2022review}. 

\emph{Spatial GCNs} on the other hand, unlike spectral methods, operate directly on the graph’s node and edge features, aggregating information from neighboring nodes to capture local dependencies, while avoiding the computationally expensive recalculation of the entire graph. Hence, spatial GCNs are particularly suitable for fault location,
given that it is a task that primarily relies
on local network structure.
Lakouraj et al.~\cite{mansourlakouraj2021application} reduce the computational complexity of spectral GCNs by using a linearized spectral filter, which can be viewed as an approximate transition toward spatial graph convolution.
Ma et al.~\cite{ma2024method} explicitly approximate spectral GCNs to spatial GCNs, 
applying the general
framework proposed by Kipf and Welling~\cite{kipf2016semi} 
to the fault location task. Another approach to incorporate the spatial structure of the feeder is~\cite{teixeira2021fault}, 
who uses Gated Graph Neural Networks (GGNNs) for fault location,  extending standard GNNs to iteratively update the hidden states of nodes through successive iterations. Zhang et al.~\cite{zhang2025fault} proposed the GDIA-GCL framework to handle fault location in distribution networks with sparse measurement infrastructure. This two-stage approach first uses a Graph Dirichlet Imputation Algorithm (GDIA) to estimate missing bus measurements, followed by a GCN model trained with Graph Contrastive Learning (GCL) to better distinguish fault signatures under extreme data scarcity.

The GNN methods discussed so far solely focus on spatial relationships, and thus do not cater specifically for the temporal dynamics inherent in fault events.
To address this shortcoming, Nguyen et al.~\cite{nguyen2022one} proposed a hybrid model that combines a 1D-CNN and a GCN for fault diagnosis (i.e., to detect, localize, and classify the faults), where the 1D-CNN is used to capture temporal patterns and then a GCN~\cite{kipf2016semi} model extracts spatial correlations. Later, in~\cite{nguyen2023spatial}, they improved this approach by adding Long Short-Term Memory (LSTM) networks, which are well suited for modeling sequential dependencies in time series data. Later, Wang et al.~\cite{wang2025fault} introduced a temporal GCN that divides time-series data into patches and uses multi-scale fusion strategies to better exploit temporal features. Together, these approaches illustrate how temporal and spatial information can be jointly addressed for fault diagnosis in power systems.

\begin{figure*}[t]
    \centering
\includegraphics[width=\textwidth]{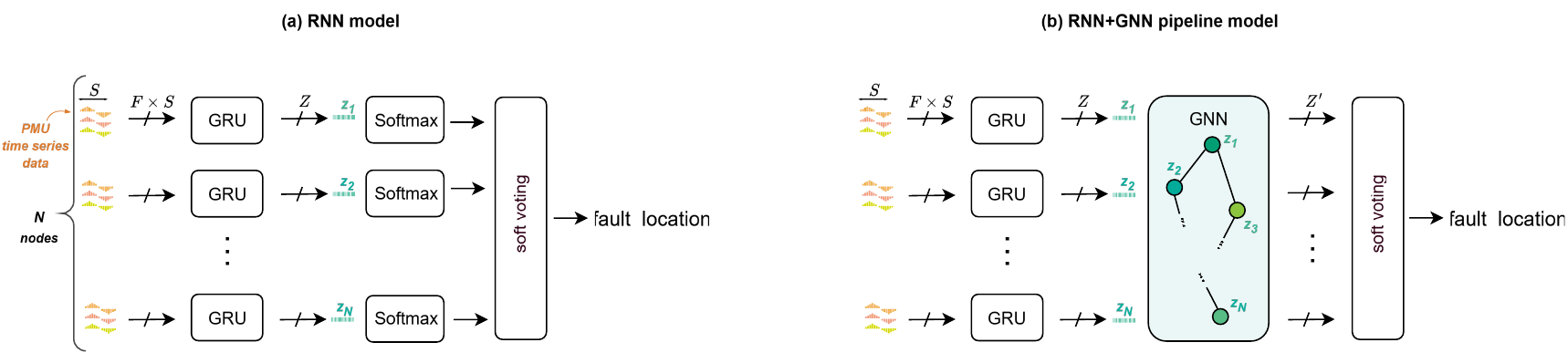}
    \Description{Model Architectures}
    \caption{
    Model architecture of the (a)~shared GRU for local classification per node, aggregated via soft voting (summing node probabilities and taking the max), (b)~STGNN pipeline, where GRU embeddings of all nodes are processed by GNN layers with message passing, dropout, and batch normalization, with each variant using its respective architecture (see \cref{subsec:sfe_gnn}), followed by soft voting.
    $N$: number of nodes, $F$: number of features (\eg phase voltages), $S$: sequence length (\ie number of timesteps in the considered measurement window),
    $Z$: GRU hidden state dimension (and hence output), $Z'$: GNN output dimension.}
    \label{fig:rgnn_architecture}
\end{figure*}

In terms of GNN model architectures, a limitation of GCNs~\cite{kipf2016semi} is that they rely on a fixed, degree-based normalization of the graph to aggregate the information from neighbors.
This dependence can reduce their adaptability when the graph structure changes~\cite{hamilton2017inductive, velickovic2018graph}, e.g., when the feeder topology is changed to switches opening/closing.
Hence, GCNs may be less suited to practical fault location scenarios, where network configurations can change.
Thus, alternatively, Graph Attention Networks (GATs)~\cite{velickovic2018graph} have been proposed: e.g.,
Wang et al.~\cite{wang2024enhanced} leverage an attention mechanism to weigh the importance of neighboring nodes when aggregating information for fault location.
By learning these node-specific weights, GATs reduce their reliance on a fixed graph structure and, in return, focus more on local connections.
To not only capture the spatial feeder structure, but also leverage temporal aspects, Ngo et al.~\cite{ngo2025deep} further extend GAT models by incorporating a 1D-CNN to capture temporal information, using both voltage and branch current inputs to detect, localize and classify faults. 
Still, these works using GATs leave some remaining challenges unaddressed.
First of all, the proposed solutions assume near-complete visibility of all 3-phase bus measurements, and do not account for 1- or 2-phase lines, thus failing to reflect realistic feeder conditions with limited measurement infrastructure deployment.
Furthermore, they did not explore the potential benefits of considering asymmetric attention mechanisms, i.e., where information might flow differently along the connection (detailed in~\cref{sec:GATv2}).  

Alternatively, GraphSAGE architectures have been explored to enhance the scalability and robustness of fault localization.
GraphSAGE shifts the learning process from global graph embeddings to localized neighborhood signatures by performing graph subsampling, and using local aggregation functions~\cite{hamilton2017inductive}.
For instance, GSAGE-GAT~\cite{Wei2024gsagegat} combines GraphSAGE’s local aggregation layers with attention-based weighting to improve the model's expressiveness across complex feeder sections.
Furthermore, to address challenges regarding measurement noise, Fan et al.~\cite{Fan2024VGAEGSage} integrated a Variational Graph Auto-Encoder (VGAE) with GraphSAGE. However, existing GraphSAGE-based approaches often (i)~neglect the temporal nature of fault waveforms and (ii)~do not sufficiently address the challenge of partial observability, where sparse sensor placement limits the available feature set at each node.

In summary, while GNNs have demonstrated significant potential for fault location, current research often relies on datasets with limited variety and primarily focuses on dense sensor placement on three-phase buses. Furthermore, these studies typically maintain a 1-to-1 mapping between the GNN graph and the physical feeder without exploring alternative graph representations that can better reflect partial observability, and there is no rigorous performance comparison across various GNN architectures.

\section{Methodology}
\label{sec:method}
To address the gaps of the current fault location studies surveyed above, we design an STGNN pipeline, including two previously unexplored variants, specifically tailored for distribution network fault location and propose a graph construction algorithm that explicitly accounts for the inherent partial observability of the grids. 
The subsequent subsections describe STGNN components, graph construction algorithm, and, in particular, discuss the GNN models in detail.

\subsection{Proposed STGNN framework}
\label{sec:proposed_framework}
As outlined in \cref{sec:related}, previous research has explored RNNs~\cite{yu2019intelligent} to capture the temporal patterns and GNNs~\cite{chen2020fault} to extract spatial patterns in isolation for fault location. Hence, spatio-temporal graph neural networks (RNN+GNNs or also known as STGNNs) have been proposed as a means to jointly model temporal and spatial dependencies~\cite{nguyen2023spatial}. Our work extends this idea, by systematically comparing a wider range of different GNN architectures, particularly\,---\,for the first time, to the best of our knowledge\,---\,adopting GraphSAGE and GATv2 for distribution grid fault location.  


We thus design an STGNN pipeline, summarized in \cref{fig:rgnn_architecture}(b), that integrates temporal and spatial feature extraction. The node-level representations $\mathbf{z}_i$ produced by the GRU~( \cref{sec:TemporalFeatureExtraction}) are then processed by a GNN~( \cref{subsec:sfe_gnn}) using the given graph structure to extract spatial relationships. Finally, the GNN outputs $\mathbf{z}'_i$ are passed through a dense classifier layer to produce node-level predictions.
These predictions are aggregated at inference time via a soft voting scheme \cite{ho1994decision}, where the probabilities for each location, including a possible 'no fault' case, are summed across all nodes to produce a final prediction $\hat{y}$ while mitigating the influence of extreme node-level outputs.


For the GNN graph, we model distribution network as a graph \( G = (V, E) \), where \( V \) is the set of \( N \) nodes (buses), i.e., \( |V| = N \), and \( E \) is the set of edges (branches) interconnecting them. 
Yet, the GNN's graph topology does not necessarily need to be a 1-on-1 mapping of the feeder's bus topology (\ie \emph{full topology}). Instead, we propose a \emph{measured-only} graph construction algorithm that reflects inherent partial observability resulting from the practical deployment constraints of real-world distribution systems, by systematically mapping measured ($\upmu$PMU) nodes as graph nodes $V$.

\begin{algorithm}[bth]
\small
\caption{Graph Construction for the `measured-only' Topology}
\label{alg:graph-construction}
\KwIn{Set of $\upmu$PMUs $\mathcal{P}$ (single-, two-, three-phase), substation $O$ (\eg bus 150 in \cref{fig:IEEE123}) and physical feeder topology $\mathcal{T}$}
\KwOut{Graph $G=(V,E)$}
$V \gets \mathcal{P}$, $E \gets \emptyset$\;
\ForEach{three-phase $\upmu$PMU $v \in \mathcal{P}$ starting from the substation $O$ and following the electrical connectivity}{
    Let $N^3_v \subset \mathcal{P}$ be the three-phase $\upmu$PMUs accessible from $v$ in $\mathcal{T}$\;
    \ForEach{neighbor $w \in N^3_v$, ordered by distance to $v$}{
    Connect $v \leftrightarrow w$, unless this creates a closed loop\;
    }
}
\ForEach{single- or two-phase $\upmu$PMU $u \in \mathcal{P}$}{

Let $N^3_u, N^{1,2}_u \subset \mathcal{P}$ be the three-phase, resp.\ the single- or two-phase $\upmu$PMUs accessible from $u$ in $\mathcal{T}$\;
  \uIf{$u$ lies between two three-phase $\upmu$PMUs $\{v, w\}$ 
  in $\mathcal{T}$}{
    Connect $u \leftrightarrow v$\;
    Connect $u \leftrightarrow w$\;
 }
  \uElseIf{$|N^3_u| \ge 1$}{
    Connect $u \leftrightarrow v$, where $v \in V'$ is the $\upmu$PMU closest to $u$ (or the unique $\upmu$PMU if $|N_u| = 1$) in $\mathcal{T}$ \;
}
   \ElseIf{$|N^{1,2}_u| > 0$}{
    Connect $u \leftrightarrow m$, where $m \in N_u^s$ is closest to $u$\;
    }
}
\Return{$G=(V,E)$}\;
\end{algorithm}

Following the systematic approach outlined in our proposed algorithm \cref{alg:graph-construction}, we derive the GNN's measured-only graph by mapping the physical electrical connectivity and multi-phase characteristics of the feeder onto a set of measured nodes ($\upmu$PMUs).
Consistent with typical distribution network configurations, we assume that three-phase $\upmu$PMUs (\(v \in \mathcal{P}\)) are located along the primary feeder backbone (main distribution lines), whereas single- or two-phase $\upmu$PMUs (\(u \in \mathcal{P}\)) are deployed solely on the lateral branches. Beginning from the substation, each three-phase $\upmu$PMU \(v\) is connected to all its adjacent three-phase $\upmu$PMUs \(w \in N_v\), beginning with those in closest electrical proximity\footnote{We refer to distance as electrical proximity along the feeder topology $\mathcal{T}$ rather than physical distance between the busses.}, provided that they do not form a closed loop in the feeder topology.
Next, Single- or two-phase $\upmu$PMUs \(u\) are connected to the closest three-phase $\upmu$PMU \(v \in N_u\), unless solitary single- or two-phase $\upmu$PMU \(u\) is located between two three-phase $\upmu$PMUs \(v\) and \(w\) forming a closed loop\,---\,in such instances, \(u\) is connected to both  \(v\) and \(w\).  Lastly, connections between single- or two-phase $\upmu$PMUs \(u\) are formed only when they serve as the sole path for current flow, \ie when they have no accessible three-phase $\upmu$PMU (\(|N_u|=0\)). 

In contrast, \emph{full-topology} graph includes both measured and unmeasured buses are included as nodes, and edges are defined by directly mapping the physical line connections between buses. This not only implies that part of the GNN input data are reduced to zeros\,---\ as no meaningful measurement data exists for unmeasured buses\,---\,but also increases the graph size, which in return increase computational load. Further, assigning zero-valued features to the unmeasured buses propagate noise~\cite{rusch2023survey} across the feeder, potentially complicating the GNN model's capability to still infer the correct conclusions from measured graph nodes. In either case, the GNN graph structure is defined by the adjacency matrix \( A \in \{0, 1\} ^{N \times N} \), with $A_{u,v} = 1$ if $(u,v) \in E$ (\ie if nodes $u$ and $v$ are directly connected by an edge), and $0$ otherwise.

\subsection{Temporal Feature Extraction -- Recurrent Neural Networks}
\label{sec:TemporalFeatureExtraction}
Fault events in distribution networks inherently form time series data. RNNs are widely used to process such temporal dependencies~\cite{cao2025fault}, with LSTM and GRU cells being well-known variants.
While LSTMs are designed to preserve long-term dependencies, fault patterns in power systems are typically confined to relatively short time windows.
Given the simpler architecture and higher computational efficiency of GRUs~\cite{cahuantzi2023comparison}, we chose to use GRU cells in our RNN+GNN models to extract temporal features from the measurements. From each single measured node \( v \in V \), a fixed time window of measurements (\ie per-phase RMS voltages, representing the root-mean-square value of the AC voltage) is fed into the GRU, which produces representations that are then input to the GNN, as illustrated in \cref{fig:rgnn_architecture}.


\subsection{Spatial Feature Extraction --  Graph Neural Networks}
\label{subsec:sfe_gnn}
In a GNN framework, each node \( v \in V \) is associated with a feature vector  \( H_v \in \mathbb{R}^{d} \), forming the node feature matrix  \( H \in \mathbb{R}^{N \times d} \), where the $i$-th row corresponds to the features of the $i$-th node in $V$. 
In GNNs in general, also edges can have their own set of features (e.g., reflecting geographical distance), which then typically modulate the influence of neighboring nodes. Yet, here we only use node features.

Graph learning aims to map node and edge features of a graph $G$ to target outputs (in our case this will be a fault location), formally represented as:
\begin{equation}
    \hat{y} = f(G; \theta),
\end{equation}  
where $\theta$ denotes the model parameters, and $\hat{y}$ represents the predicted output.
GNNs learn node representations by iteratively aggregating information from neighboring nodes across layers:
\begin{equation}
    H^{(k+1)} = f(H^{(k)}, A; \theta), \label{eq:gnn-general}
\end{equation}
where $H^{(k)} \in \mathbb{R}^{N \times d_k}$ represents the node representation matrix at the output of layer $k$. This matrix is formed by row vectors $h^{(k)}_v$, with each vector representing a node $v \in V$ with a $d_k$-dimensional feature vector at layer $k$. This iterative aggregation is known as a message passing framework, in which each node $v$ updates its representation 
by aggregating the features from its neighbors $u \in \mathcal{N}(v)$, with the neighborhood $\mathcal{N}(v)$ including the node $v$ itself via self-loops, to ensure its features are preserved during aggregation (\ie $v \in \mathcal{N}(v)$).
GNN architectures vary in how message passing is performed, using different aggregation rules and non-linear activation functions \(\phi(\cdot)\) applied after aggregation to compute \( f \) in \cref{eq:gnn-general}. 

Our study involves three GNN variants:
\subsubsection{Graph Convolutional Networks (GCN)} 
In a GCN, the node representations are updated by aggregating features from neighboring nodes using a normalized adjacency matrix $\hat{A}$~\cite{kipf2016semi}. This update can be expressed as: 
\begin{align}
    H^{(k+1)} &= \phi \left( \hat{A} \> H^{(k)} \> W \right), &W \in \mathbb{R}^{d_k \times d_k}
    \label{eq:eq6} 
    \\
    \text{with } \hat{A} &= D^{-1/2} \> A' \>  D^{-1/2}, &\hat{A} \in \mathbb{R}^{N \times N}. \label{eq:eq5}
\end{align}
\noindent
Here $W$ denotes a learnable weight matrix, and $d_k$ is the feature dimension. 
The modified adjacency matrix $A' = A + I$ adds self-loops by including the identity matrix $I$. This is the matrix representation of the augmented neighborhood $v \in \mathcal{N}(v)$. 
The degree matrix $D$ normalizes the adjacency, weighting the contributions of neighboring nodes based on their connectivity in the graph.
Thus, $\hat{A}$ amounts to fixed degree-normalized connectivity, that is used to aggregate features from neighboring nodes and further weighted by $W$.

\subsubsection{GraphSAGE (Graph Sample and Aggregation)}
In GraphSAGE, node’s representation is updated using a \emph{learned} aggregation function over its neighbors~\cite{hamilton2017inductive}, rather than relying on a fixed normalization that depends on the given topology, as the degree-normalized adjacency $\hat{A}$ that GCN uses. 
This implies that GraphSAGE can more flexibly
accommodate structural changes in the graphs, such as variations in node neighborhoods, at inference.
The updated feature vector for node $v$ at layer $k+1$ is given by:\footnote{The original GraphSAGE model alternatively proposes a sampled subset $\mathcal{N}'(v) \subset \mathcal{N}(v)$ rather than the full set of neighbors. However, as the number of measured nodes is limited and full-batch training remains computationally efficient, we consider GraphSAGE without subsampling (see \cref{sec:model-training}).} 
\begin{equation}
    h_v^{(k+1)} = \phi \left( W  \cdot\, \text{concat}\left( h_v^{(k)}, \text{AGG}\left( \{ h_u^{(k)} | u \in \mathcal{N}(v) \} \right) \right) \right).
\end{equation}
In this update, $W$ is a learnable weight matrix, $\text{concat}$ is the concatenation operation, and $\text{AGG}$ represents a neighborhood aggregation function. Note that, unlike the standard GraphSAGE~\cite{hamilton2017inductive}, we include the node itself in the neighborhood set (i.e. $v \in \mathcal{N}(v)$) to reinforce the node's features alongside the explicit concatenation. 

\subsubsection{Improved Graph Attention Network (GATv2)} 
\label{sec:GATv2}
GATv2 introduces an attention-based message passing mechanism that adaptively weighs the contributions of neighboring nodes~\cite{brody2022attentive}. The representation of a node $v$ at layer $k+1$ is computed as:
\begin{equation}
     h_v^{(k+1)} = \phi\left( \sum_{u \in \mathcal{N}(v)} \alpha_{vu} \> W h_u^{(k)} \right).
\end{equation}
Here, \( \alpha_{vu} \in \mathbb{R} \) represents the attention coefficient between nodes $v$ and $u$, determining the relative importance of the features of node $u$ for updating node $v$, and is calculated as:
\begin{equation}
    \alpha_{vu} = \frac{\exp\left( a^T \cdot \text{LeakyReLU}(W_1 \> h_v^{(k)} + W_2 \> h_u^{(k)}) \right)}{\sum_{j \in \mathcal{N}(v)} \exp\left( a^T \cdot \text{LeakyReLU}(W_1 \> h_v^{(k)} + W_2 \> h_j^{(k)}) \right)}, \label{eq:GATv2}
\end{equation}
where $a$ is a learnable attention vector and $\text{LeakyReLU}$ denotes a nonlinear activation. 

The earlier Graph Attention Network (GAT) proposed by Velickovic et al.~\cite{velickovic2018graph} computed attention as a linear combination of features ($W \> h_v || \> W h_u$), which applies the same linear transformation $W$ to both nodes before concatenation, and the nonlinearity is applied after the attention is computed. In contrast, GATv2 provides
\begin{enumerate*}[(i)]
\item more flexibility by using separate learnable weights
($W_1$ and $W_2$) that allow for asymmetric neighbor contributions, and \item greater expressiveness by applying the nonlinearity (LeakyRELU) before the attention vector ($a$) multiplication.
\end{enumerate*}

\section{Experimental Setup}
\label{sec:exp}

\subsection{Fault Types}
\label{sec:fault-types}
Power system faults are broadly divided into two categories: open-circuit and short-circuit faults~\cite{gururajapathy2017fault}. \textit{Open-circuit} faults arise when a conductor breaks, interrupting the normal flow of current, whereas \textit{short-circuit} faults occur due to unintended contact between conductors or between a conductor and ground, leading to excessive current flow. Among these, short-circuit faults receive the most attention in fault location studies, given their higher frequency and significant impact on system stability. Short-circuit faults can be further classified as asymmetrical faults (line-to-ground (LG), line-to-line (LL), and double line-to-ground (LLG)) and symmetrical faults such as three-phase faults \cite{grainger1994power}.

In practice, these broad categories translate into 11 distinct fault types: three LG faults (AG, BG, CG), three LL faults (AB, BC, CA), three LLG faults (ABG, BCG, CAG), and two symmetrical types (three-phase, ABC; and three-phase-to-ground, ABCG). In our study, we consider all 11 fault types to capture the diversity of fault scenarios in practice and enable a thorough assessment of fault location.

\subsection{Simulation Setup and Data Collection}
\label{sec:simulation-setup}
The performance of the proposed data-driven models depends on access to high-quality and diverse datasets. Since fault data in distribution networks are scarce\,---\,due to the infrequent occurrence of faults and inherent partial observability occurring from limited measurement infrastructure\,---\,we generate synthetic data\footnote{Generating synthetic data is standard practice in distribution grid fault studies as real-world datasets are scarce and typically restricted due to security, and proprietary concerns~\cite{nguyen2023spatial, chen2020fault}.} via simulations in OpenDSS~\cite{epri_opendss} using the PyDSS interface~\cite{pydss}. Our experiments are carried out on the IEEE~123-bus feeder, a widely adopted benchmark in fault diagnosis studies~\cite{nguyen2022one, nguyen2023spatial, chen2020fault}, which operates at a frequency of \SI{60}{\hertz} and a nominal voltage of \SI{4.16}{\kilo\volt}.

\begin{figure}[t]
\centering
\includegraphics[width=\columnwidth]{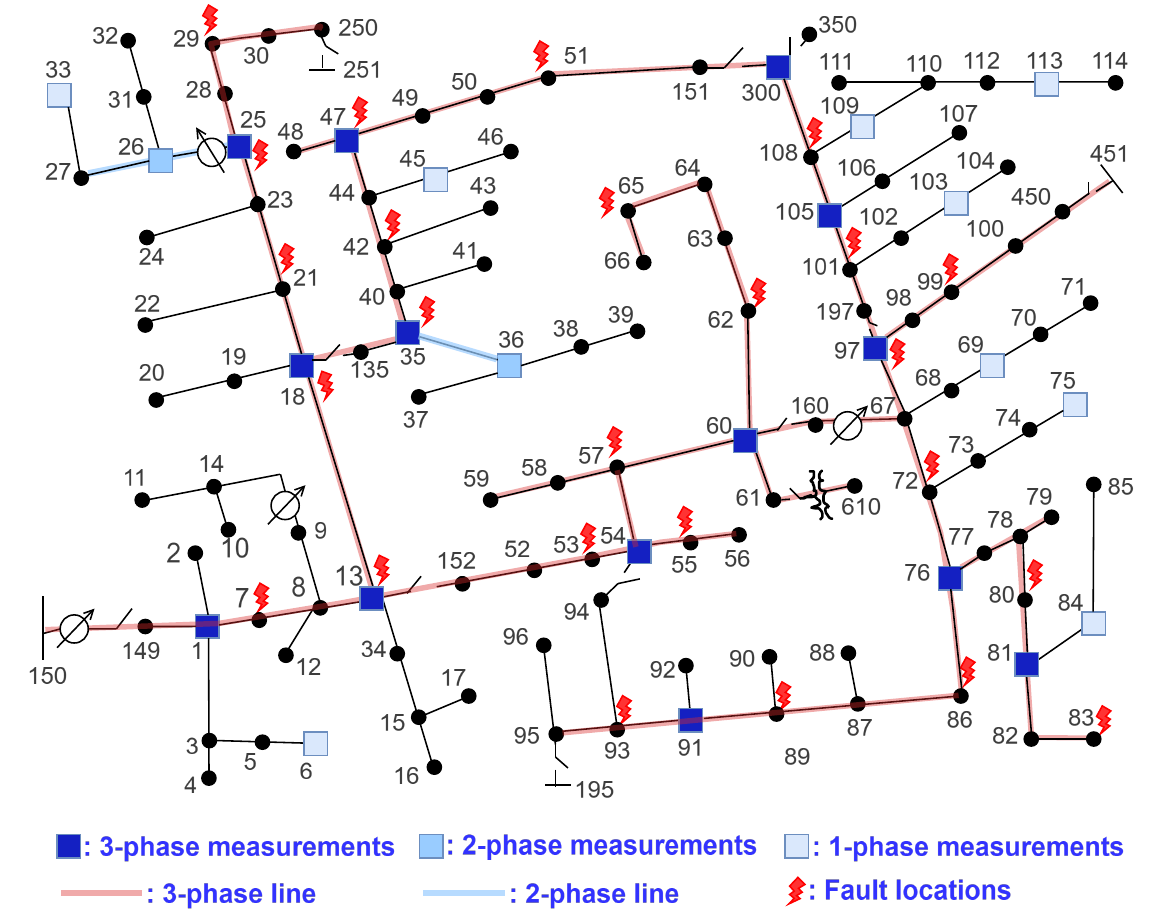}
\Description{IEEE~123-node feeder with fault locations and voltage measurements.}
\caption{IEEE~123-node feeder with fault and $\upmu$PMU locations.}
\label{fig:IEEE123}
\end{figure}

Short-circuit Fault durations in distribution systems typically range from \SI{20}{\milli\second} to \SI{50}{\milli\second}, reflecting a typical primary protection clearing time; in this study, we fix the duration at \SI{20}{\milli\second} for consistency. As illustrated in \cref{fig:IEEE123}, faults are dynamically injected at 25~different locations, each covering one of the 11~fault classes, while measurements are recorded by $N=\text{25}$ idealized micro-phasor measurement units ($\upmu$PMUs) (which partly coincide with some of the considered failure locations). Measurement data are sampled at \SI{1}{\milli\second} resolution, such that each fault event is represented by a sequence of 20~samples, enabling detailed fault profile analysis.

For each of the 25 fault scenarios, we run 100 different simulations, from which we record the three-phase RMS voltage magnitudes ($V_1, V_2, V_3$) at each of the 25 $\upmu$PMU locations.
To create sufficient diversity among such 100 runs per fault case, we vary both the bus loads and the resistance on the faults (between lines or from line to ground).
For the bus loads, to represent a wide spectrum of operating conditions, we scale the load given by the standard IEEE~123-bus configuration: for each of the physical buses, we independently multiply its original load with a factor $L \sim \mathcal{U}(0.5, 1.3)$. 
Here, the lower bound ($0.5$) represents off-peak conditions (\eg nigh-time when residential and commercial activity is at its lowest), while the upper bound ($1.3$) represents peak demand periods (\eg mid-afternoon when industrial consumption is at its highest).
The resistance for the fault in a particular scenario is picked from $R_f \in \{0.1, 1, 10\}~\Omega$, representing near-bolted, low and moderate fault resistance respectively.

From the $\upmu$PMU measurement data streams, we record a \SI{60}{\milli\second} window, 
starting \SI{40}{\milli\second} before the fault, 
from which we thus extract 40 overlapping sliding windows of $S=\text{20}$ timesteps (\SI{20}{\milli\second}).
Each sliding window serves as one input sample to our classification models (GRU or STGNN). Half of the windows correspond to the fault-free condition, while the other half capture fault events. In total, this yields \num{2.5} million windows, with 50\% non-fault cases and 2\% for each fault location. 
\Cref{table:dataset_info} summarizes our data generation configurations. These specifications apply independently to each feeder configuration considered in this study (detailed in \cref{sec:model-training}), including the switch-reconfigured state used for resilience testing.
Finally, the dataset is randomly partitioned into 70\% training, 15\% validation, and 15\% test sets.
To preserve temporal consistency across $\upmu$PMUs, sliding windows from the same time step are kept together during splitting.

As features, voltage is selected over current, as it is more commonly available through existing infrastructure without the need for additional sensors; furthermore, voltage provides a more stable signal across the feeder for fault analysis~\cite{nguyen2023spatial}. We specifically use RMS values instead of raw instantaneous waveforms, as this is consistent with the data that $\upmu$PMUs output in practice, while it also reduces communication overhead~\cite{ieee2011pmu}.

\begin{table}[t]
\caption{Dataset Information} 
\centering
\resizebox{\columnwidth}{!}{
\begin{tabular}{lccc} 
\toprule
\toprule
\textbf{Elements} & \multicolumn{2}{l}{\textbf{Value}} & \textbf{Number} \\
\midrule
Fault type & \multicolumn{2}{l}{AG, BG, CG, AB, BC, CA, ABG, BCG, CAG, ABC, ABCG} & 11 \\
\midrule
Fault resistance ($\Omega$) & \multicolumn{2}{l}{0.1, 1.0, 10} & 3 \\
\midrule
Fault position (Buses) & \multicolumn{2}{p{6cm}}{7, 13, 18, 21, 25, 29, 35, 42, 47, 51, 53, 55, 57, 62, 65, 72, 80, 83, 86, 89, 93, 97, 99, 101, 108} & 25 \\
\midrule
$\upmu$PMU location & \multicolumn{2}{p{6cm}}{1, 6, 13, 18, 25, 26, 33, 35, 36, 45, 47, 54, 60, 69, 75, 76, 81, 84, 91, 97, 103, 105, 109, 113, 300} & 25 \\
\midrule
Load Multiplier (p.u.) & \multicolumn{2}{l}{Uniformly sampled from [0.5, 1.3]}  & 1000 \\
\bottomrule
\toprule
\multicolumn{4}{c}{\textbf{Dataset Distribution}} \\
\midrule
\textbf{Category} & \textbf{Train Set} & \textbf{Validation Set} & \textbf{Test Set} \\
\midrule
Fault Cases &  875,000 & 187,500 & 187,500 \\
\midrule
\replacetext{Non-Fault}{No Fault} Cases &  875,000 & 187,500 & 187,500 \\
\midrule
Total Cases & 1,750,000 & 375,000 & 375,000 \\
\bottomrule
\end{tabular}
} 
\label{table:dataset_info}
\end{table}

\begin{figure*}[t]
    \centering
    \begin{subfigure}[b]{0.45\textwidth}
        \centering      
        \includegraphics[width=\textwidth]{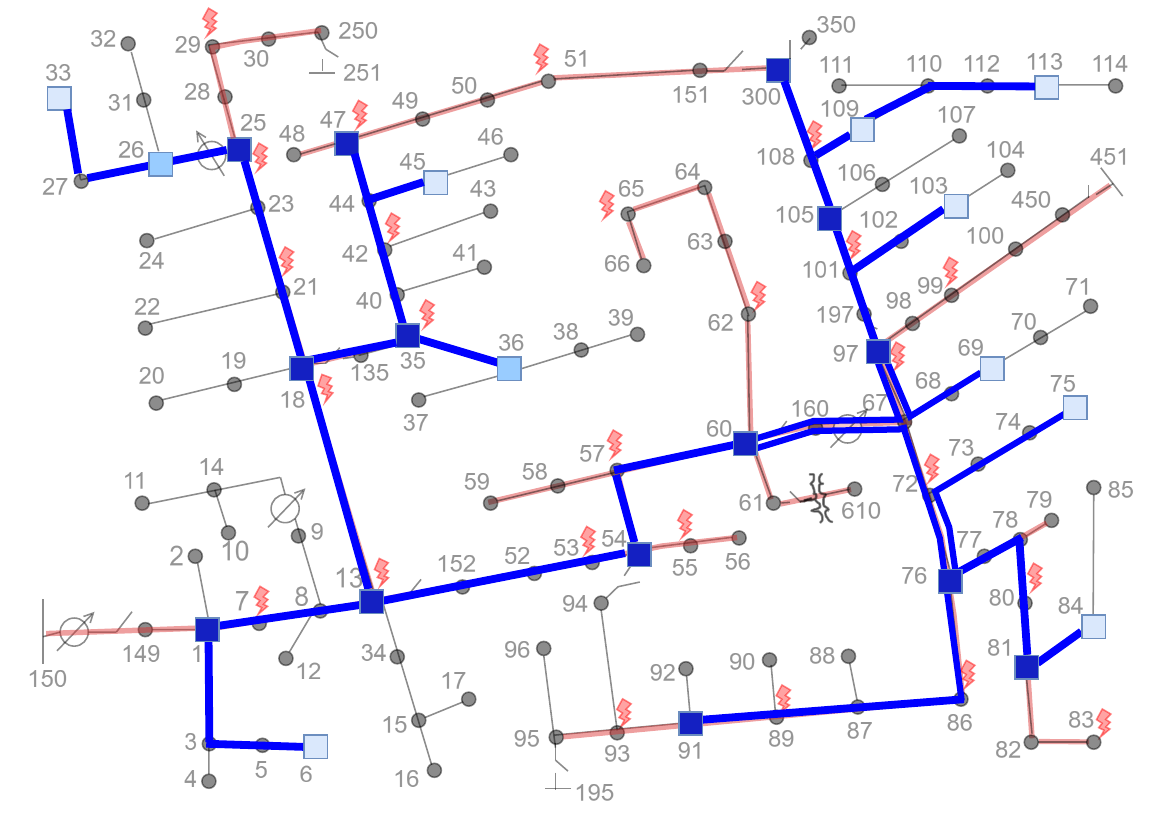}
        \caption{Default configuration (Blue)}
        \label{fig:default_config}
    \end{subfigure}
    \hspace{0.01\textwidth}
    \vrule
    \hspace{0.015\textwidth}
    \begin{subfigure}[b]{0.45\textwidth}
        \centering
    
\includegraphics[width=\textwidth]{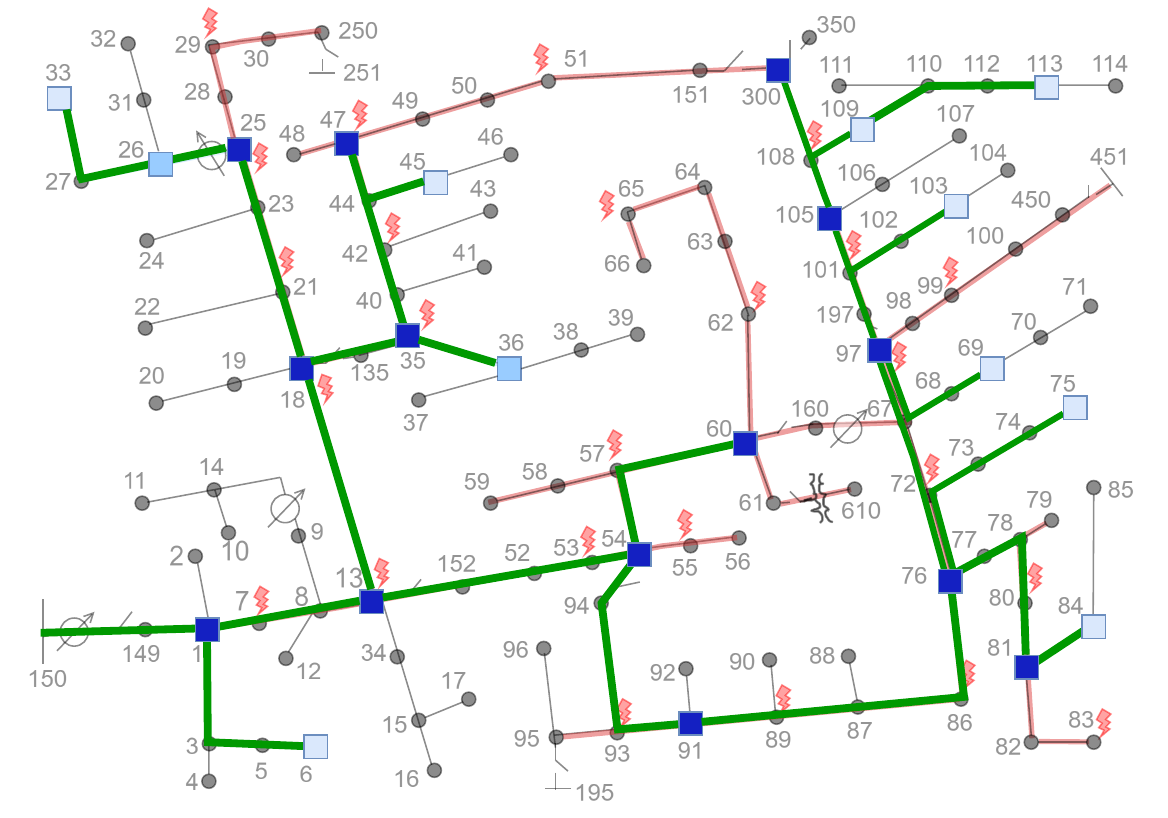}
        \caption{Lateral-redirected configuration (Green)}
        \label{fig:green_config}
    \end{subfigure}
    \caption{IEEE~123-node feeder measured-only graph representations, drawn to resemble the feeder layout for easier visual correspondence. The default configuration (a) is shown on the left (blue), while the lateral-redirected configuration (b) on the right (green) is achieved via switching operations, resulting in subtle voltage variations across the feeder.}
    \Description{Side-by-side graph representations of an IEEE~123-node feeder: default configuration on the left, the lateral-redirected green configuration on the right.}
    \label{fig:measured-only_graph}
\end{figure*}

\subsection{Model Training and Evaluation}
\label{sec:model-training}

Fault classification is formulated as a multi-class problem with \num{26}~labels (\num{25}~fault locations, and a no-fault label) and optimized using the cross-entropy loss function. During STGNN training, classification is applied at the node level, producing a separate prediction for each node. At inference time, graph-level predictions are obtained by aggregating the node-level outputs using a soft voting scheme, as described in \cref{sec:proposed_framework}.
As a baseline to assess the performance of GNN-based solutions, we consider a pure RNN solution: an aggregated \emph{GRU} model~(\cref{fig:rgnn_architecture}(a)) applied independently to each $\upmu$PMU and node-level predictions are then aggregated using the same soft voting process. 

We implemented all models in PyTorch, using the PyTorch Geometric (PyG) library~\cite{fey2019fast} for GNN models, on a device equipped with an Intel Core i5-9500 CPU (3.00 GHz) and 64\,GB of RAM. We normalize all features $(V_1, V_2, V_3)$ using Z-score normalization, such that each feature has a mean of \num{0} and a standard deviation of~\num{1}. To ensure a fair comparison, both the proposed STGNN framework and the GRU baseline use an identical temporal extraction layer consisting of GRU cells with a hidden state size of \num{128}. In the STGNN architecture, this is followed by GNN layer with a hidden dimension of \num{64}. For the GraphSAGE-based models (RGSAGE), we use full-batch training with either mean or max pooling as the aggregation function~\cite{hamilton2017inductive}. While GraphSAGE is widely known for enhancing scalability via neighborhood sampling (mini-batch setting) in massive graphs, full-batch training is highly feasible in our study since we only have a modest number of measured nodes. Finally, for all models, these latent representations are passed through a dense layer to produce a \num{26}-dimensional logits for each node.
Rectified Linear Units (ReLU) are used as activation functions within the network. Softmax is applied to the output logits at inference time to obtain class probabilities, which are then used for soft voting aggregation. During training, PyTorch’s cross-entropy loss~\cite{paszke2019pytorch} internally applies log-softmax to the logits for numerical stability. To mitigate overfitting, we apply dropout \num{0.35} and batch normalization; for GAT models, we additionally use an attention dropout rate of \num{0.3} and use four attention heads. Models are trained using the AdamW optimizer~\cite{llugsi2021comparison} with a batch size of \num{32}. AdamW extends the Adam algorithm by decoupling weight decay from the gradient-based updates. We set the weight decay to \num{1e-4} and the learning rate to \num{0.001}. The models are trained untill convergence. Performance is evaluated using Macro F1 score on the test set.

Measurement devices such as $\upmu$PMUs are costly, and typically not installed at each and every bus in the distribution grid, but only a subset of them. To reflect this partial observability, we construct GNN graph using our \emph{measured-only} graph construction algorithm (detailed in \cref{sec:proposed_framework}) that uses only the $N=25$ measured buses as nodes. \cref{fig:default_config} illustrates the resulting reduced \emph{measured-only} graphs obtained for the IEEE 123-bus feeder. 
Then, to systematically assess the impact of our \emph{measured-only} approach, we contrast it  against commonly used 1-on-1 mapping \emph{full topology} baseline ($N=128$, including the $123$ standard buses plus $5$ auxiliary buses for regulators, transformer secondaries, and switches) to assess the impact of graph representation on fault location. In this case zero feature values are used for the unmeasured buses, and their outputs are masked during loss computation and final prediction.

To further evaluate our \emph{measured-only} graph strategy and the
models under more challenging operating conditions, we perform switch reconfiguration\,---\,which we refer to as the \textit{green configuration}\,---\,specifically by opening the tie-switch between buses 60 and 160 and closing the switch between buses 54 and 94 (shown in \cref{fig:green_config}). The green configuration introduces structural variations in the network and reroutes power flow through lateral branches, leading to weaker fault signatures as we observe more subtle voltage variations across the measured nodes. We train and evaluate models on their respective configurations.

\begin{figure*}[t]
    \centering
\includegraphics[width=1\textwidth]{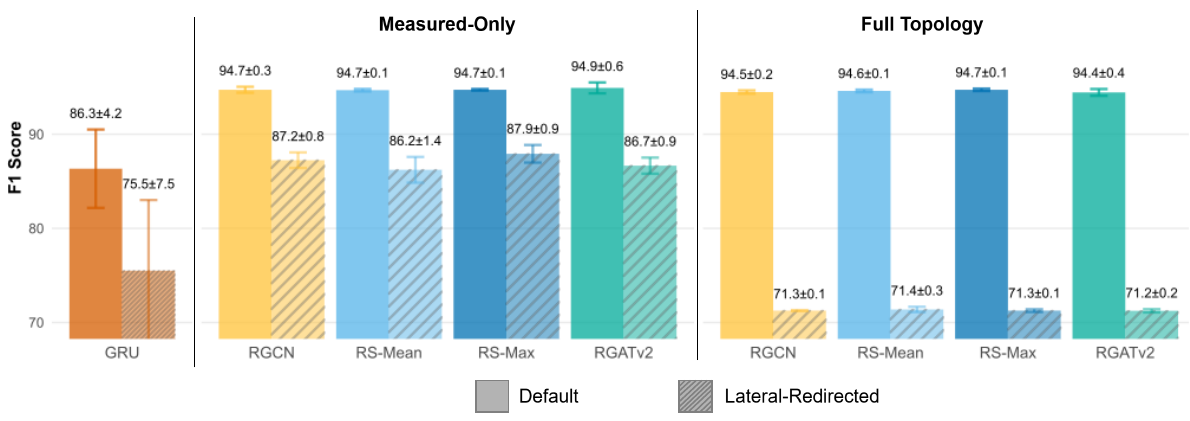}

    \Description{Comparative fault location results on default and green configurations.}
    \caption{F1 scores for fault location comparing the \emph{default} (solid bars) and lateral-redirected \emph{green} (hatched bars) configurations, where RGSAGE models are denoted as RS. The results are grouped by topology: GRU baseline (left), \emph{measured-only} (middle), and \emph{full topology} (right). Error bars represent the 90\% confidence intervals across models trained with different random seeds. }    
    \label{fig:combined_results}
\end{figure*}


\section{Results and Discussion}
\label{sec:res}
We now present a quantitative performance analysis of the GRU and STGNN models on the IEEE~123-bus system across two distinct feeder configurations: the default configuration and the \emph{green} configuration characterized by more subtle voltage variations and weaker fault signals (see \cref{fig:green_config}).
We first compare the fault location performance across the RNN baseline and various STGNN models (\cref{sec:rnn-vs-rgnn}). 
Next, we examine the benefit of our proposed \emph{measured-only} topology over the conventional \emph{full-topology} graph strategy (\cref{sec:logical-topology}). 
In both cases, we will also investigate the influence of the feeder topology on the performance achieved by the respective models and graph construction methods. 

\subsection{RNN and STGNN Model Comparison}
\label{sec:rnn-vs-rgnn}
To evaluate the performance of different architectures for fault location under sparse $\upmu$PMU coverage, we perform a systematic benchmarking study by comparing our proposed RNN+GNN models (RGSAGE and RGAT) against a state-of-the-art graph model (RGCN) and a non-GNN baseline (GRU).
Specifically, we investigate 
\begin{enumerate*}[(i)]
\item whether explicit spatial modeling via GNNs provides superior performance over a purely temporal GRU baseline, and
\item which GNN architecture performs best under sparse measurement conditions.
\end{enumerate*}
Note that we here discuss the various GNN-based models' performance using the \emph{measured-only} topology graphs.

We first examine the fault location performance of the RNN-only and RNN+GNN models for the \emph{default configuration}. As \cref{fig:combined_results} shows, the \emph{RNN-only model} using GRU architecture as discussed previously, only reaches F1\,$\simeq$\,\num{0.86}. This is consistent with the performance of RNN-based fault location on the IEEE~123-bus system as observed in~\cite{nguyen2023spatial},
who consider a similar dataset and data processing conditions.\footnote{Nguyen et al.~\cite{nguyen2023spatial} obtain 85.3 fault location F1 for their LSTM model.} 
Note that our result is obtained by using the soft-voting aggregation of individual nodes' predictions (as discussed in \cref{subsec:sfe_gnn}): for GRU this boosts the node-level predictions with an increase of roughly \num{0.10} in F1 score.

In contrast, the \emph{GNN-based models} achieve a substantial improvement over the RNN baseline, with the F1 score rising from $\simeq$,\num{0.86} to $\simeq$,\num{0.95}.
Notably, for these models, the soft-voting step adds only $\sim$\,\num{0.01} to the F1 score.
This better performance of GNN-based models, along with the comparably smaller impact of aggregation, is what we intuitively expect from GNNs' ability to process topological relationships (\ie structure and connectivity of the graph), allowing them to effectively combine distributed inputs into a global view through message passing.

Interestingly, as shown in \cref{fig:combined_results}, all GNN-based variants yield statistically similar results (F1 ranging from $94.7 \pm 0.3$ to $94.9 \pm 0.6$).
This suggests that for the \emph{default} IEEE~123-bus feeder configuration, the spatial dependencies are sufficiently captured even by relatively simple GNN architectures such as RGCN.
This again aligns with the findings in~\cite{nguyen2023spatial}, where\,---\,under the similar dataset and processing conditions mentioned earlier\,---\,transitioning from an RNN to a GNN yielded a significant performance jump from $\sim$\,\num{0.85} to $\sim$\,\num{0.92}.

Now, when we consider more challenging grid conditions\,---\,which, best to our knowledge, related works did not study yet\,---\,such as \emph{green} configuration shown in \cref{fig:green_config}, we obtain qualitatively similar observations.
In that \emph{green} configuration, the redirected power flow through lateral branches leads to weaker fault signatures. This reduces the magnitude of voltage deviations across the measured nodes, making the distinction between fault-induced transients and nominal load profiles more subtle.
Hence, as shown in \cref{fig:combined_results}, we observe a performance decrease across all architectures. The GRU baseline’s F1 score drops to $75.5$, with its confidence interval widening from $4.4$ to $7.5$, indicating increased instability under weaker fault signal conditions.
In contrast, the GNN models maintain a much tighter performance cluster between F1 \num{86} and \num{88}.

These results demonstrates the GNNs' superior performance and robustness, both across randomness in the training process (i.e., for different random seeds) and, perhaps more importantly, over different physical grid feeder configurations.
Indeed, the GRU exhibits lower overall performance and a wide confidence interval that increases considerably under the more challenging \emph{green} configuration, while the GNN variants maintain significantly higher mean $F1$ scores and much narrower error bars.
This superiority of GNN-based solutions over pure RNN ones is consistent with intuition and confirms earlier assessments in literature.

Regarding our comprehensive benchmarking of multiple GNN architecture variants, ultimately, we do not find a definitive superiority of one GNN architecture over the others.
Mostly, their performance seems on par, for both the \emph{default} as the more challenging \emph{green} configuration. 
Still, from the numerical results, we do observe that the relative ranking of models shifts under more challenging fault signal conditions. While RGATv2 nominally leads in the default case, RGSAGE-Max ($87.9 \pm 0.9$) performs best in the green configuration. 
This potentially suggests that when fault signals are less pronounced (hence the relative importance of individual nodes is harder for the model to identify), the attention mechanism as adopted in RGATv2 does not provide a distinct benefit over simpler aggregation; instead, max-pooling is more effective at capturing the subtle, peak signals necessary for fault location in green configuration. 

Since GNN architectures were originally designed to process larger and more complex networks (\eg social network graphs), that all considered GNN models are equally successful, in hindsight may be unsurprising.
Yet, we hypothesize that advanced GNN models may offer more noticeable benefits when we complicate the fault location problem, \eg  when evaluating its generalization to a priori unseen feeder configurations (and thus different logical GNN graphs), 
\eg to maintain robust performance of pre-trained models to cases where additional $\upmu$PMUs are installed at new locations and/or the distribution feeder's topology changes because of changed switch configurations.

\subsection{Logical Graph Topology Comparison}
\label{sec:logical-topology}
Now that we have established the superiority of GNN-based models perform best, we investigate the impact of the choice of GNN graph topology construction on fault location performance. More specifically, we compare STGNN models trained on the traditionally used \emph{full-topology} graphs (where unmeasured nodes are included in the graph topology, but masked with zeros) with those using our proposed  \emph{measured-only} graphs to evaluate their relative performance and computational complexity.

For the \emph{default configuration}, 
we note that there is virtually no performance difference between the two topology choices, which both achieve F1\,$\simeq$\,\num{95}, and thus our proposed \emph{measured-only} topology does not offer a performance benefit.
However, for the \emph{green} configuration\,---\, where fault signals are less pronounced compared to default configuration\,---\,
per\-formance for the STGNN models based on the \emph{full-topology} graph drops significantly, down to F1\,$\simeq$\,\num{71}, 
whereas our \emph{measured-only} approach loses way less and still achieves F1\,$\simeq$\,\numrange{86}{88}.
The observed performance drop for the \emph{full topology} models in the green configuration, at first may seem a little counter-intuitive: despite the models now being informed of the complete feeder topology, classification performance decreases.
Such decrease can be explained by the message passing mechanism of GNNs: the non-$\upmu$PMU nodes, that are now also part of the \emph{full-topology} graph, obviously have no measurement data (i.e., its local features have zero values), which can dilute information from measured buses during aggregation.
While this effect is seemingly limited for the default grid configuration, where fault signals are more pronounced, the dilution effect proves rather critical in green configuration.

In addition to providing superior performance also under weaker fault signal conditions, using our \emph{measured-only} graph models has another notable advantage: reduced computational requirements for training the classification models.
Indeed, training when using \emph{full-topology} graphs takes roughly \num{6}$\times$ longer than for \emph{measured-only} graphs, as summarized in \cref{tab:time_complexity}.\footnote{Note that we report recorded training run times for the challenging \emph{green} grid configuration. Given that the model parameters and number of training samples are not dependent on the grid configuration, and thus training run times are virtually the same for the \emph{default} grid configuration.} 

Although all the GNN-based models share the same asymptotic complexity, i.e., their training time scales linearly with the number of neighbors $k$ and feature dimension $d$ (\ie $O(kd)$), RGATv2 is the most computationally intensive for both graph topology choices, due to the per-edge attention weight calculations (see \cref{subsec:sfe_gnn}).
Theoretically, RGSAGE-Max is more expensive than RGSAGE-Mean, because maximum aggregation requires element-wise comparison and argmax tracking logic across neighbors, whereas mean aggregation relies on more hardware-efficient summation and averaging.
Given the sparse connectivity of the IEEE~123-bus feeder\,---\,where most nodes have only 1--2 neighbors\,---\,the theoretical difference between mean- and max-pooling computation times disappears into the background noise of the training environment and seed-specific convergence rates.
We conclude that RGCN remains the most efficient and stable model ($363.17\,\pm\,5.70$\,min), benefiting from pre-computed adjacency normalization and avoiding the concatenation overhead found in GraphSAGE (see \cref{subsec:sfe_gnn}). 

\begin{table}[ht]
\centering
\caption{
Run times for training the STGNN models 
(mean $\pm$ standard deviation, across \num{5} training runs with different random initialization).}
\label{tab:time_complexity}
\begin{tabular}{lcc}
\toprule
\textbf{Model} & \multicolumn{2}{c}{\textbf{Training Time (min)}} \\
\cmidrule(lr){2-3}
 & Measured-only & Full-topology \\
\midrule
RGATv2       & 65.07\,$\pm$\,1.18 & 391.34\,$\pm$\,19.62 \\
RGSAGE-Max   & 62.22\,$\pm$\,1.14 & 372.38\,$\pm$\,22.13 \\
RGSAGE-Mean  & 61.40\,$\pm$\,0.76 & 373.33\,$\pm$\,19.82\\
RGCN         & 61.56\,$\pm$\,0.70 & 363.17\,$\pm$\,5.70\phantom{0}\\
\bottomrule
\end{tabular}
\label{Table:timeComplexity}
\end{table}

In summary, our results demonstrate that RNN+GNN (STGNN) models are highly effective for distribution grid fault location.
First, GNN-based architectures demonstrate a clear and consistent superiority over the RNN-only baseline (i.e., the GRU model).
In the \emph{default configuration}, RGNNs offer an \num{0.9} F1 improvement (reaching F1\,$\simeq$\,\num{0.95}), while maintaining a \,$\simeq$\,\num{0.12} lead for the \emph{green} grid configuration.
Beyond F1 scores, STGNNs prove significantly more robust; their confidence intervals remain narrow in all cases, whereas the GRU exhibits high uncertainty (that further widens from \num{4.2} to \num{7.5} under weaker fault signal conditions, \ie the \emph{green} grid configuration).
Second, we find that, while all tested GNN variants are effective, their relative performance is context-dependent. RGATv2 nominally leads in the default case where node importance is more distinct, but RGSAGE-Max (F1\,=\,$87.9\,\pm\,0.89$) proves most effective in the green grid configuration. This suggests that max-pooling acts as a robust peak-signal extractor, preserving subtle fault signatures that would otherwise be diluted by the weighted averaging of attention mechanisms.

Furthermore, our proposed \emph{measured-only} topology approach demonstrates critical advantages over the conventional \emph{full-topology} approach.
While both strategies perform similarly in commonly studied grid conditions, our \emph{measured-only} approach proves significantly more resilient to challenging grid parameters: under conditions of weaker fault signals, it maintains an F1 between \num{86} and \num{88}, whereas the \emph{full-topology} performance collapses to $\sim$~\num{0.71}, due to noise propagation from zero-masked nodes.
Coupled with a 6-fold reduction in model training time, our findings indicate that a \emph{measured-only} GNN framework provides a more robust, stable, and computationally efficient solution for real-world distribution networks.

\section{Conclusion and Future Work}
\label{sec:conclusion}
The aim of this paper was to comprehensively and systematically 
\begin{enumerate*}[(i)]
\item compare recent GNN architectures for the task of distribution network fault location, for which we constructed a large-scale dataset of three-phase RMS voltage measurements with sparse $\upmu$PMU placement,\label{it:analyze-comprehensively}
and
\item investigate and systematically compare graph-forming strategies beyond the complete or near-complete feeder topology, for which we introduced a reduced, \emph{measured-only} graph topology that directly reflects the inherent partial observability of the distribution grids. \label{it:investigate_graph}
\end{enumerate*}
For \cref{it:analyze-comprehensively}, we proposed previously unexplored RNN+GNN models, namely RGSAGE and an improved RGAT variant, and benchmarked them against the state-of-the-art RGCN as well as RNN-only baselines.
For \cref{it:investigate_graph}, we are, to the best of our knowledge, the first to systematically compare representative graph-forming strategies: the conventionally adopted \emph{full-topology} representation including all (and thus also unmeasured) buses, and our reduced \emph{measured-only} topology.
To assess the robustness of our findings, we evaluated these models across two distinct scenarios: the \emph{default configuration} representing the standard IEEE~123-bus feeder, and a \emph{green configuration} obtained through switch reconfigurations, where lateral-redirected path result in much smaller and more subtle voltage changes under fault conditions (see \cref{sec:model-training}).

Our study shows that GNN-based models consistently outperform RNN-only baselines for fault location, achieving F1\,$\simeq$\,\num{0.95} for the default IEEE 123-bus grid configuration and \,$\simeq$\,\num{0.86}-\num{0.88} under more challenging fault signal conditions of green configuration across all GNN variants. Furthermore, GNN-based  approaches demonstrate superior stability in their predictions: while the RNN baseline's uncertainty widens significantly under challenging grid conditions (with confidence intervals expanding from \num{4.2} to \num{7.5} percentage points F1), the STGNNs maintain narrow confidence intervals ($\leq\num{1.4}$ percentage points F1), demonstrating STGNNs' robustness even when faced with weaker fault signals.

Regarding our proposed \emph{measured-only} graph topology approach, we find that it:
\begin{enumerate*}[(a)]
\item provides significant resilience to challenging grid parameters, maintaining an F1 between \num{0.86}-\num{0.88} where the \emph{full-topology} approach collapses to $\simeq$~\num{0.71} due to noise propagation, and
\item reduces model training times by a factor of \num{6}$\times$.
\end{enumerate*}
This suggests that the measured-only representation is a more practical choice, offering substantial efficiency gains while maintaining stable performance.
Besides more extensive analysis of the effect of graph topology choice, further GNN refinements that may boost fault location performance include 
\begin{enumerate*}[(a)]
\item incorporating edge features such as physical line distances, which may provide richer structural information, and
\item exploring deeper GNN architectures to better understand their impact.
\end{enumerate*}
The study of these latter items is left for future work.

Looking ahead, we also plan to further explore what benefits the newly explored GraphSAGE and RGATv2 models may bring.
Even though in our above study they did not achieve notable improvements upon state-of-the-art GCN models for the studied configurations, we hypothesize that GraphSAGE and RGATv2 could offer stronger generalization and transferability of learned models across different feeder configurations and topologies.
Specifically, the inductive nature of GraphSAGE makes it a prime candidate for transferring learned fault-location inference to entirely new feeders, while the attention mechanism of GATv2 may better adapt to complex, non-linear dependencies in diverse switch configurations.
Our future work will therefore extend this study towards exploring generalization capabilities of the proposed GNN-based models, as well as studying their ability to solve additional fault diagnosis tasks, e.g., as fault type classification.

\begin{acks}
Research reported in this publication was supported by VITO grant number VITO\_UGENT\_PhD\_2301 and
partially funded by the Flemish Government (AI Research Program).
\end{acks}



\end{document}